\documentclass{article}
\usepackage{times}
\usepackage{soul}
\usepackage{url}
\usepackage[hidelinks]{hyperref}
\usepackage[utf8]{inputenc}
\usepackage[small]{caption}
\usepackage{graphicx}
\usepackage{amsmath}
\usepackage{amsthm}
\usepackage{amsfonts}
\usepackage{booktabs}
\usepackage{algorithm}
\usepackage{algorithmic}
\usepackage{authblk}
\usepackage[switch]{lineno}

\newtheorem{mydef}{Definition}
\DeclareMathOperator*{\argmin}{arg\,min}
\title{A dataset-free approach for self-supervised learning of 3D reflectional symmetries
}

\author[1]{Isaac Aguirre}
\author[1]{Ivan Sipiran}
\author[1]{Gabriel Montañana}
\affil[1]{Department of Computer Science, University of Chile}
\date{}
\begin{document}

\maketitle

\begin{abstract}
    In this paper, we explore a self-supervised model that learns to detect the symmetry of a single object without requiring a dataset—relying solely on the input object itself. We hypothesize that the symmetry of an object can be determined by its intrinsic features, eliminating the need for large datasets during training. Additionally, we design a self-supervised learning strategy that removes the necessity of ground truth labels. These two key elements make our approach both effective and efficient, addressing the prohibitive costs associated with constructing large, labeled datasets for this task. The novelty of our method lies in computing features for each point on the object based on the idea that symmetric points should exhibit similar visual appearances. To achieve this, we leverage features extracted from a foundational image model to compute a visual descriptor for the points. This approach equips the point cloud with visual features that facilitate the optimization of our self-supervised model. Experimental results demonstrate that our method surpasses the state-of-the-art models trained on large datasets. Furthermore, our model is more efficient, effective, and operates with minimal computational and data resources.

\end{abstract}

\section{Introduction}

One of the most important structural properties of shapes is symmetry, which allows them to remain unchanged after certain transformations. Symmetry occurs naturally in living organisms, such as the wings of insects and birds, inanimate substances, such as snowflakes, and can even be introduced artificially into man-made structures, such as ancient pyramids.

\begin{figure*}[ht!]
    \centering
  \includegraphics[width=\textwidth]{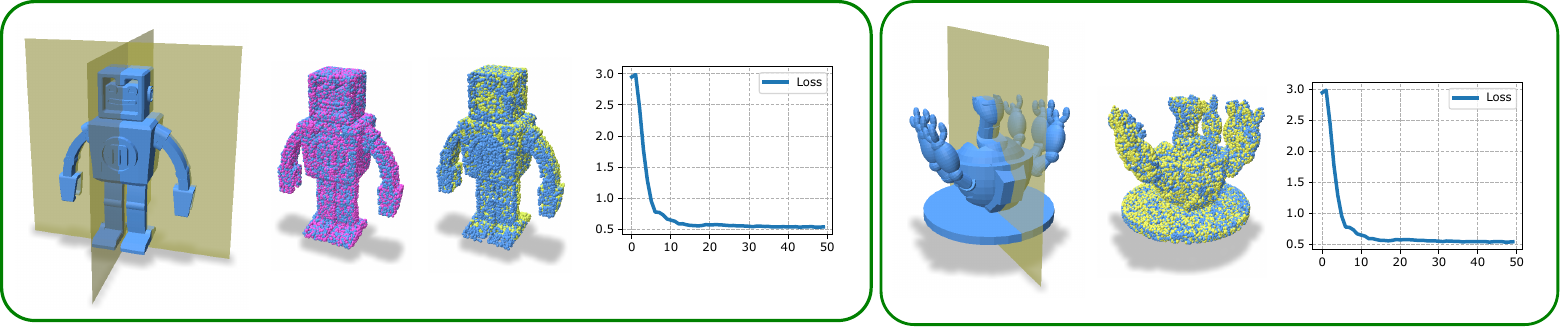}
  \caption{The figure shows the detected symmetries along with the overlapping point clouds (original and symmetric) for two examples of the Thingi10K dataset. Our method converges in very few iterations, which makes it very efficient in practice.}
  \label{fig:teaser}
\end{figure*}

Symmetry is important in problem solving, because it compresses the information required to represent an entity. Typically, the computational representations of objects do not include symmetrical information, which can be used to save space by repurposing symmetrical parts and storing the rest separately. Moreover, manually providing symmetric information for an object can be difficult and tedious, in addition to developing software for tags. This raises the motivation to construct an automatic tool to analyze the representation of symmetrical information.

The availability of symmetric information can aid many high-level tasks in computer graphics and vision. Some use cases are in CAD applications \cite{Li2008,Chang2008,Jiang2013a,Li2014c} where the storage space of 3D parts can be reduced, in Computational Modeling to enhance the features of 3D modeled faces \cite{Liao2012}, and in archaeology \cite{Son2013,Xia2013,Harary2014b,Li2014b,Sipiran2014,Mansouri2015,Mavridis2015} where symmetries can be used to reconstruct or improve the representation of cultural heritage objects discovered in damaged states.

Classical methods for detecting symmetry make geometrical assumptions that are difficult to formulate in an algorithm that generally contains many parameters, with tasks such as finding symmetrical points that are already complex. Other methods that work relatively well by analyzing the differential geometry in surfaces can take hours to run and may even fail to find candidates if the assumptions are not correctly made. Recently, well-conditioned neural network methods have been shown to be proficient at finding solutions in significantly shorter times, which is an advantage when integrated into larger applications. However, one of their problems is their reliance on datasets of varied objects. When objects are too varied and their symmetries highly complex, the model may learn a common structure but fail to find specific symmetries, making it unable to output shapes with specific structures.

This work endeavors to apply learning-based techniques to solve the problem of symmetry detection, with a focus on exploiting self-prior \cite{point2mesh}, which has shown promise as a way to train a neural network for an input object without a collection of other objects. We believe that a natural way to solve the problem of detecting the symmetries of an object is to analyze only its structure, without relying on the existence of other samples for that purpose. This form of learning is related to the concept of deep internal learning, which seeks to take advantage of an entity's own information, in contrast to external learning, where one seeks to find common patterns in a dataset.

There are many advantages to a method that learns from a single object over methods that learn from a dataset in the symmetry detection problem, even more so if the method does not require supervision of the standard form. First, it is not necessary to collect a large amount of data or label them. Second, the method does not learn common patterns from a dataset but rather specializes in solving the problem for a given sample, making this approach ideal for specific cases. Third, this approach is particularly effective for deployment in real-world shape analysis applications, as it is efficient and does not rely on pretrained models.

In this paper, we propose a simple but effective methodology that uses the concept of self-priors to detect symmetries in 3D shapes. The main idea of our research is that, by using a carefully trained neural network with a loss function that comes from the very definition of the formal concept of symmetry, it is possible to detect the symmetric structures of an input object in a self-supervised manner (Fig. ~\ref{fig:teaser} shows examples of our method applied on shapes from Thingi10K dataset~\cite{thingi10k}).

The remainder of this paper is organized as follows. Section~\ref{Sec:related} describes relevant related works. Section~\ref{Sec:background} introduces the background required to elaborate our proposal. Section~\ref{Sec:overview} provides an overview of the proposed method. Section~\ref{Sec:Features} presents the method to compute visual features for a point cloud. Section~\ref{Sec:reflectional} provides details on the detection of reflectional symmetries. Finally, Section~\ref{Sec:Conclusions} concludes this paper.

\section{Related Work}
\label{Sec:related}
There are many methods that address the problem of symmetry detection from a geometric perspective. These approaches rely on finding points or regions that mutually correspond according to symmetry. In general, this approach relies on solving a self-matching problem in a 3D shape \cite{Mitra2010,Raviv2010a,Raviv2010b,Berner2011,Wang2011,Liu2012,Shehu2014,Tevs2014,Yoshiyasu2014}. For example, Liu et al. \cite{Liu2012} formulated the symmetry detection problem by finding stationary points in a symmetric transformation. To find some evidence of symmetric points over the surface, the Blended Intrinsic Maps method~\cite{Kim2011} was applied. The method detects potential keypoints in the 3D shape and computes a metric of geodesic distances between the keypoints and the remaining points on the surface.  Blended maps were computed using geodesic distances as constraints. The zero-level curves of the blended maps were considered as the potential curve axes of symmetry.
				
\subsection{Learning-based methods}
Learning-based methods formulate the symmetry detection problem as a learning task, where a neural network processes a 3D shape to output a symmetrical representation \cite{Bu2015,Ji2019}.

Gao et al. \cite{Gao2020} proposed a neural network that predicts planar reflective and rotational symmetries for objects. Input shapes are converted into voxel representations, and the method predicts up to three symmetries, which are validated during post-processing. Notably, this approach does not require labeled data for symmetry detection, classifying it as an unsupervised method. However, the detection capacity is limited to at most three symmetries. Experiments conducted on the ShapeNet dataset \cite{Chang2015} utilized known object alignments to generate ground truth for evaluation.

Shi et al. \cite{Shi2020} introduced a multi-task framework to detect reflective and rotational symmetries using RGB-D images (RGB images augmented with depth information). They highlight a critical limitation of traditional learning approaches, which often memorize likely symmetries for specific objects rather than generalizing to unseen cases. To mitigate overfitting, their model learns to find symmetries and symmetric correspondences directly. The ground truth is derived automatically using an optimization-based symmetry detection method, enabling the model to simulate accurate symmetry predictions. While RGB input provides essential context for the task, the approach is not purely geometric. Instead, point-wise features are computed using PointNet \cite{pointnet}, integrating geometric and visual information.

Recently, Li et al. \cite{e3sym2023} introduced an unsupervised model for detecting reflectional symmetries by leveraging $E(3)$ invariance within a point cloud encoder. The proposed framework consists of three key components: an initial module that extracts invariant features from the input point cloud, a feature-matching step to identify candidate symmetric correspondences, and a differentiable grouping strategy to generate the final set of detected symmetries. 

\subsection{Self-prior for geometric problems}
 
\paragraph{Point2Mesh}
The models described by Shi et al.~\cite{Shi2020} and Gao et al.~\cite{Gao2020} require datasets for training, for which an alternative is considered, using the idea that a model may learn to self-adjust geometric properties based only on the observed object~\cite{point2mesh,selfsampling}. The self-prior concept also sees an earlier appearance applied to images in the  work of Ulyanov et al. \cite{deepimageprior}, with many useful applications such as denoising, super-resolution, and autocompletion. 
		
A prior represents problem-specific knowledge that can be incorporated into a model beyond the training data, such as through architectural choices, assumptions about data distribution, or proximity in embedded space. A self-prior is a type of prior learned automatically from a single input, where the model's structure alone represents the input without reliance on external training data.

Point2Mesh \cite{point2mesh} addresses the reconstruction of 3D models by iteratively refining an initial mesh to fit an input point cloud using a deep neural network. This method leverages self-similarity in the input shapes by utilizing shared weights in convolutional layers, enabling feature learning independent of spatial location while mitigating the effects of outliers. An encoder-decoder network maps partial inputs to complete shapes, with a loss function based on Chamfer Distance or Earth Mover’s Distance to compare the decoded output with the ground truth.

\paragraph{Self-Sampling}
Point cloud consolidation is a problem in which the goal is to remove noise, outliers, or add new points to a given point cloud input. The neural network model developed using Self-Sampling for Neural Point Consolidation \cite{selfsampling} learns only from the input cloud. Generally, it is difficult to scan sharp regions in a model, even though they may serve as distinctive parts of the object, making it a candidate for a consolidation criterion and motivating its generation by adding points through computerized means.
		
Sampling routines were established to separate the model into positive and negative points, where positive refers to points that fit the desired consolidation criteria. In this study, a loss function was developed to compare the Chamfer Distance between a sample of the input cloud consisting of negative points to one of the positive points, with the ultimate goal of making a network learn a mapping from negative to positive.

\section{Background}
\label{Sec:background}
This section summarizes the main definitions required to understand our proposal. As our research is devoted to analyzing global symmetries in rigid bodies, we focus on the definitions for this specific case.

\begin{mydef}
Symmetry is a nontrivial transformation $T$ such that when we apply this transformation to a 3D object $\mathcal{O}$, the result is the same object. Formally, $T$ is a symmetry if $T(\mathcal{O}) = \mathcal{O}$.
\end{mydef}

By non-trivial, we mean that $T$ is not an identity. We would also like to point out that the previous definition is strict concerning required equality. In practice, $\mathcal{O}$ is a discrete representation of the continuous surface of a 3D object; hence, sampling introduces errors that should be considered when defining the symmetry. Therefore, we introduce an alternative approximate definition.

\begin{mydef}
Let $\mathfrak{S}$ be the space of 3D objects and let $d: \mathfrak{S} \times \mathfrak{S} \rightarrow \mathbb{R}$ be a function that measures the congruence between two 3D objects. A transformation $T$ is an $\alpha$-approximated symmetry of object $\mathcal{O}$ if $d(\mathcal{O}, T(\mathcal{O})) \leq \alpha$.
\end{mydef}

Note that this definition is more general than exact symmetry, which can be recovered when $\alpha=0$. Note also that these definitions hold only when symmetry applies to the overall object, commonly known as \emph{global symmetry}.

In terms of the transformations of interest, we restrict $T$ to transformations in the rigid world. In this scenario, $T$ is extrinsic because it preserves the Euclidean metric in the 3D space. Given two points $p_i, p_j \in \mathcal{O}$, transformation $T$ is extrinsic if $\|p_i - p_j\|_2 = \|T(p_i) - T(p_j)\|_2$. That is, $T$ preserves the pairwise Euclidean distances between points in $\mathcal{O}$. Examples of extrinsic transformations are rotations, reflections, and translations, for which we focus our attention on reflections. The reflection symmetry can be characterized by the normal of the reflective plane and a point on the plane. 

Because we assume that the objects have global symmetry, we can consider the centroid of the object to be the fixed point of the symmetric transformation without loss of generality, as the position of objects can be normalized. Therefore, the only values to be predicted are the vectors that characterize the respective planes (specifically, the three floating-point numbers).

\section{Overview}
\label{Sec:overview}
We model the symmetric transformation T using a neural network with parameters $\Phi$ and the symmetry detection task for a single object $\mathcal{O}$ as the following optimization problem:

\begin{figure*}[ht!]
\centering
\includegraphics[width=\textwidth]{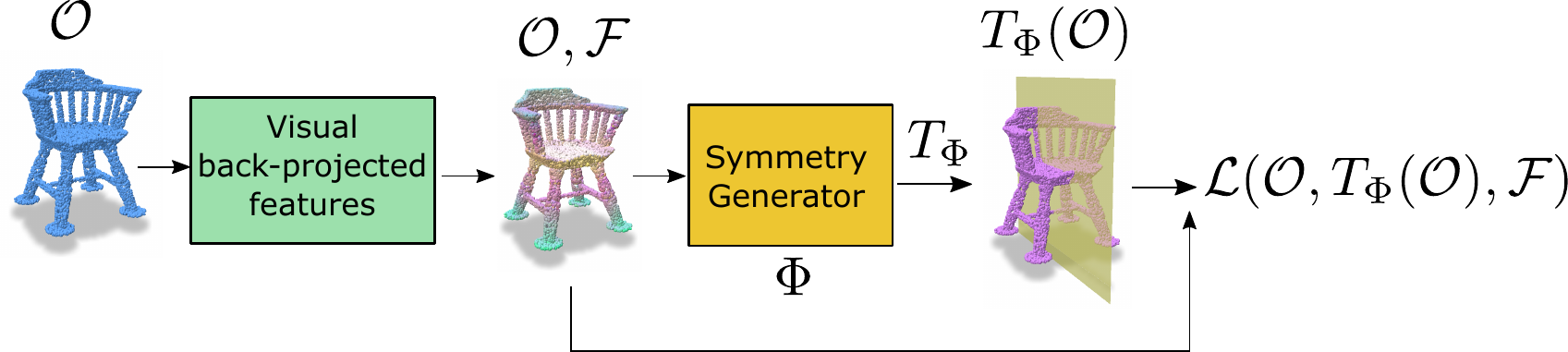}
\caption{A symmetry generator looks at the input point cloud $\mathcal{O}$ and predicts a transformation $T_{\Phi}$ to apply to $\mathcal{O}$. A modified Chamfer Loss is carried out between $\mathcal{O}$, $T_{\Phi}(\mathcal{O})$ and the features $\mathcal{F}$. The model is trained to determine $T_\Phi$ such that $\mathcal{O}$ and $T_\Phi(\mathcal{O})$ correspond symmetrically. The expected output of the model is $T_\Phi$.}
\label{Fig:proposal1}
\end{figure*}

\begin{equation}
\label{Eq:OptimizationProblem}
    T^* = \argmin_{T_{\Phi}} \mbox{  } d(\mathcal{O}, T_{\Phi}(\mathcal{O}))
\end{equation}

\noindent where $d(\cdot,\cdot)$ is a function that measures the congruence between two 3D objects. Figure~\ref{Fig:proposal1} presents an overview of the proposed method.

Note that our optimization problem comes from the definition of the approximated symmetry. Solving the optimization problem in Eq.~\ref{Eq:OptimizationProblem} is equivalent to determining the $\alpha-$approximated symmetry with the lowest $\alpha$ as possible.

Our formulation involves three key ingredients. First, we preferred point clouds as representations of 3D objects. Point clouds provide a concise and compact representation of a 3D object and facilitate the representation of an arbitrary topology for the input 3D object. Hence, henceforth, $\mathcal{O} \in \mathbb{R}^{n\times 3}$ is a 3D point cloud of $n$ points. Second, the choice of architecture for a neural network depends on the input representation. Fortunately, there are effective architectures for processing 3D point clouds, such as PointNet~\cite{pointnet}, PointNet++~\cite{pointnet++}, and DGCNN~\cite{dgcnn}. Moreover, we show that the simplest architecture already allows us to obtain good results in symmetry detection, demonstrating the efficiency and effectiveness of our proposal. Finally, the choice of congruence function leads to good self-supervised signals for the downstream task. We compute features for 3D points that contain rich visual information, facilitating the optimization process. Additionally, we propose an extended version of the Chamfer distance that integrates these features into the minimization of the problem defined in Equation~\ref{Eq:OptimizationProblem}.

\section{Point Cloud Features}
\label{Sec:Features}
Our strategy is to equip a 3D object with salient features that preserve certain invariance to symmetries, thereby aiding in the detection of its structures. While existing approaches capture features at the geometric level, we believe that visual information serves as an ideal complement for this task, especially given the availability of foundational models that adapt effectively to new tasks.

The back-projection of features extracted from images to a 3D object has been successfully applied to tasks such as 3D keypoint detection~\cite{Wimmer2024} and registration~\cite{Caraffa2024}. These methods share a common methodology: First, a set of viewpoints around the object is selected. Second, images of the object are rendered from these viewpoints. Third, a foundational model is used to extract features from the rendered images. Finally, the image features are back-projected onto the 3D object.

A key factor in determining the effectiveness of the extracted features lies in the choice of viewpoints. Generally, viewpoints are uniformly distributed on a sphere enclosing the object, with points arranged equiangularly in both horizontal and vertical directions. While this strategy suffices in most cases, we observed that under arbitrary object rotations, the resulting features are not always symmetric. We attribute this limitation to the challenges foundational models (specifically, transformer-based encoders) face in handling visual information under arbitrary rotations. To address this, we propose a novel method for capturing 3D object information that ensures improved invariance to arbitrary rotations.

Our first decision is to adopt Fibonacci sampling~\cite{Fibonacci2009} on the sphere to distribute viewpoints. Fibonacci sampling provides a better distribution of points compared to traditional equiangular approaches, which we hypothesize contributes to better rotational invariance.

Our second decision is to maximize the amount of visual information the encoder obtains from the object. For each viewpoint, we render an image with the camera's up vector aligned to the positive Y-axis. Subsequently, we rotate the rendered image by 90, 180, and 270 degrees. As a result, for each viewpoint, we generate four views, which are then fed into the encoder to extract features. Notably, rotating the rendered image achieves the same effect as rotating the camera at the corresponding viewpoint. However, this approach offers computational efficiency, as rendering is performed only once per viewpoint.

Given an input object in the form of a triangular mesh, we use Dinov2~\cite{oquab2024} to compute features from rendered images. This model outputs patch-level features, so we apply bilinear interpolation to compute pixel-level features. Since we have access to rendering information, we can determine the pixel corresponding to each vertex of the object, and assign the associated feature to that vertex. As a vertex may be visible in multiple images, the final feature for each vertex is computed as the average of all features from the images where the vertex appears.

Subsequently, we reduce the dimensionality of the features from 384 (the feature dimension of the Dinov2 Small model) to three dimensions using Principal Component Analysis (PCA). We observed that this dimensionality reduction does not negatively impact the task of symmetry detection while significantly improving the computational efficiency of the process.

An additional consideration is that we sample points on the surface of the initial triangular mesh using barycentric coordinates. We then apply barycentric interpolation to obtain the final features for the resulting point cloud. This final processing step helps mitigate issues caused by poor triangulations or inadequate sampling of the object's surface.

\begin{figure}[ht!]
\centering
\includegraphics[width=1.0\columnwidth]{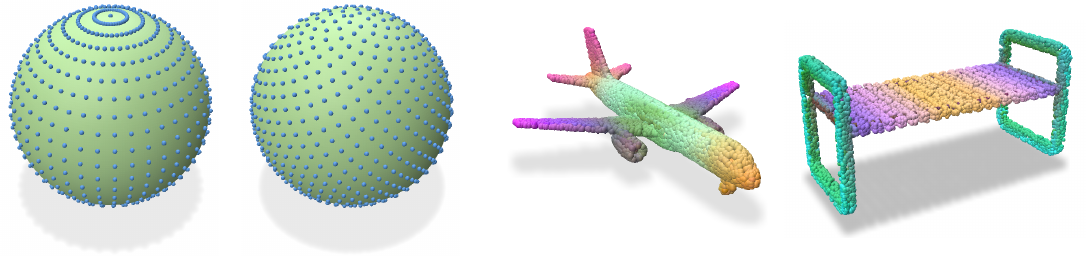}
\caption{Left: uniform and Fibonacci sampling on a sphere. Right: Plot of features as colors. Note the symmetric consistency of the features.}
\label{Fig:features}
\end{figure}

The final representation of a 3D object in our proposed method is defined as a pair $(\mathcal{O}, \mathcal{F})$, where $\mathcal{O}$ represents the 3D point cloud and $\mathcal{F}$ corresponds to the associated 3D features of the point cloud. Figure~\ref{Fig:features} illustrates the differences between uniform and Fibonacci sampling on a sphere. The figure also highlights how the features exhibit symmetric consistency.

\section{Detection of Reflectional Symmetries}
\label{Sec:reflectional}
The symmetry generator is composed of a point cloud encoder followed by a multi-head regression layer that outputs $l$ normal vectors that represent the reflective planes. Formally,

\begin{equation}
    \vec{N}_i = L_i(Encoder(\mathcal{O})), \mbox{ for } i = 1,2,\ldots, l
    \label{Eq:architecture}
\end{equation}

\noindent where $Encoder(\cdot) \in \mathbb{R}^{n\times 3} \rightarrow \mathbb{R}^{dim}$ is a point cloud neural network that outputs a $dim-$dimensional embedding for the input point cloud, $L_i \in \mathbb{R}^{dim} \rightarrow \mathbb{R}^3$ is a linear layer that outputs a normal vector. It is important to note that the encoder is shared for all prediction heads.

During the optimization, the normals $\vec{N}_i$ are normalized and subsequently transformed into a 3D transformation matrix via Householder transformation. Let $H: \mathbb{R}^3 \rightarrow \mathbb{R}^{3\times3}$ be the Householder transformation defined as

\begin{equation}
    H(\vec{N}) = \mathbb{I}_{3\times 3} - 2\vec{N}\vec{N}^T
\end{equation}

\noindent where $\mathbb{I}_{3\times 3}$ is the 3D identity transformation matrix.

One way to measure the congruence of the predicted symmetric transformation between the original object $\mathcal{O}$ and the transformed object $\mathcal{\hat{O}} = \mathcal{O} \times H(\vec{N})$ is by calculating the Chamfer distance. The Chamfer distance is defined as follows: 

\begin{align}
    CD(\mathcal{O}, \mathcal{\hat{O}}) = & \frac{1}{|\mathcal{O}|} \sum_{x \in \mathcal{O}} \min_{y \in \mathcal{\hat{O}}} \|x - y\|^2_2 \nonumber \\ 
    + & \frac{1}{|\mathcal{\hat{O}}|} \sum_{y \in \mathcal{\hat{O}}} \min_{x \in \mathcal{O}} \|x-y\|^2_2
    \label{Eq:CD}
\end{align}

We propose an extended version of the Chamfer distance that incorporates the previously computed features. This enhanced distance metric enables a more comprehensive assessment of the transformation by leveraging both geometric and feature-based information. The extended Chamfer distance is defined as follows:

\resizebox{.92\linewidth}{!}{
\begin{minipage}{\linewidth}
\begin{align}
    CD(\mathcal{O}, \mathcal{\hat{O}}, \mathcal{F}) = & \frac{1}{|\mathcal{O}|} \sum_{x \in \mathcal{O}} \min_{y \in \mathcal{\hat{O}}} \left( \|x - y\|^2_2 + \|\mathcal{F}_x-\mathcal{F}_y\|^2_2 \right) \nonumber \\ 
    + & \frac{1}{|\mathcal{\hat{O}}|} \sum_{y \in \mathcal{\hat{O}}} \min_{x \in \mathcal{O}} \left( \|x-y\|^2_2 + \|\mathcal{F}_x-\mathcal{F}_y\|^2_2\right)
\end{align}
\end{minipage}
}
\noindent where $\mathcal{F}_x$ represents the feature vector for the point $x \in \mathcal{O}$.

We use the extended Chamfer distance to define a loss term for optimization. If the neural network computes $l$ normal vectors, the loss term is defined as follows: 

\begin{equation}
    \mathcal{L}_{congruence} = \sum_{i=1}^{l} CD(\mathcal{O}, \mathcal{O}\times H(\vec{N}_i), \mathcal{F})
    \label{Eq:loss_con}
\end{equation}

\noindent where $CD$ is the extended Chamfer distance and operator $\times$ is the matrix multiplication.

In addition, one way to ensure that all normals are not equal is to impose a regularization term that encourages orthogonality of the predictions. Given the normalized vectors $\vec{N}_i$, we arrange the vector into a matrix $M = [\vec{N}_1,\vec{N}_2, \ldots, \vec{N}_l]$ of size $3\times l$, and formulate the regularization term as

\begin{equation}
    \mathcal{L}_{reg} = \||M^TM| - \mathbb{I}_{l\times l}\|
    \label{Eq:Reg}
\end{equation}

\noindent where $|\cdot|$ is the pointwise absolute value and $\|\cdot\|$ is the Frobenius norm.

Finally, the loss function for training the proposed method is formulated as $\mathcal{L} = \mathcal{L}_{congruence} + \mathcal{L}_{reg}$. Note that for cases in which one would like to detect only one symmetry, the regularization term can be discarded, leaving the optimization to depend only on the congruence term. 

\paragraph{Optimization.}
For all reported results, we conducted optimization using 100 iterations, a batch size of one, the AdamW optimizer, and a learning rate of 0.05. The encoder is a PointNet network~\cite{pointnet}. The input object initially contains 10,000 points. During training, at each iteration, we randomly sample 5,000 points. We hypothesize that this strategy facilitates faster convergence by preventing the network from becoming trapped in local minima. Additionally, we consistently enforce the network to predict three planes of symmetry. These planes are subsequently filtered during inference to determine the final result.

\paragraph{Inference.}
Once the network completes the optimization, we freeze its weights, sample 5,000 points from the input object, and predict the three planes of symmetry using the network. Finally, we discard any symmetry plane that yields an original Chamfer distance (Eq.~\ref{Eq:CD}) greater than 0.02.

\section{Experiments}
\label{Sec:Experiments}
\subsection{Dataset}
We use the ShapeNet dataset~\cite{Chang2015}, which has been previously employed by other studies ~\cite{Gao2020,e3sym2023} to evaluate the performance of symmetry detection algorithms. Since our method does not require training on a dataset, we only consider the 1,000 test objects commonly used for this task. The test objects are triangular meshes that include ground truth symmetry plane information. Notably, in this dataset, each object has at most three symmetry planes.

\subsection{Metrics}
\paragraph{Symmetry Distance Error.} This metric measures the congruence between the original input shape and its symmetrical counterpart. In practice, the symmetric reflection is applied on 1000 random points sampled on the original triangular mesh. Thus, the symmetric distance error is defined as

\begin{equation}
    SDE(O, \mathcal{S}) = \frac{1}{|O|} \sum_{o \in \O} d_{surf}(o, \mathcal{S}),
\end{equation}

\noindent where $O$ is a set of points sampled from surface $\mathcal{S}$ after the symmetric transformation. Function $d_{surf}$ is the minimum distance from a 3D point to a triangular mesh.

\begin{figure*}[ht!]
\centering
\includegraphics[width=\textwidth]{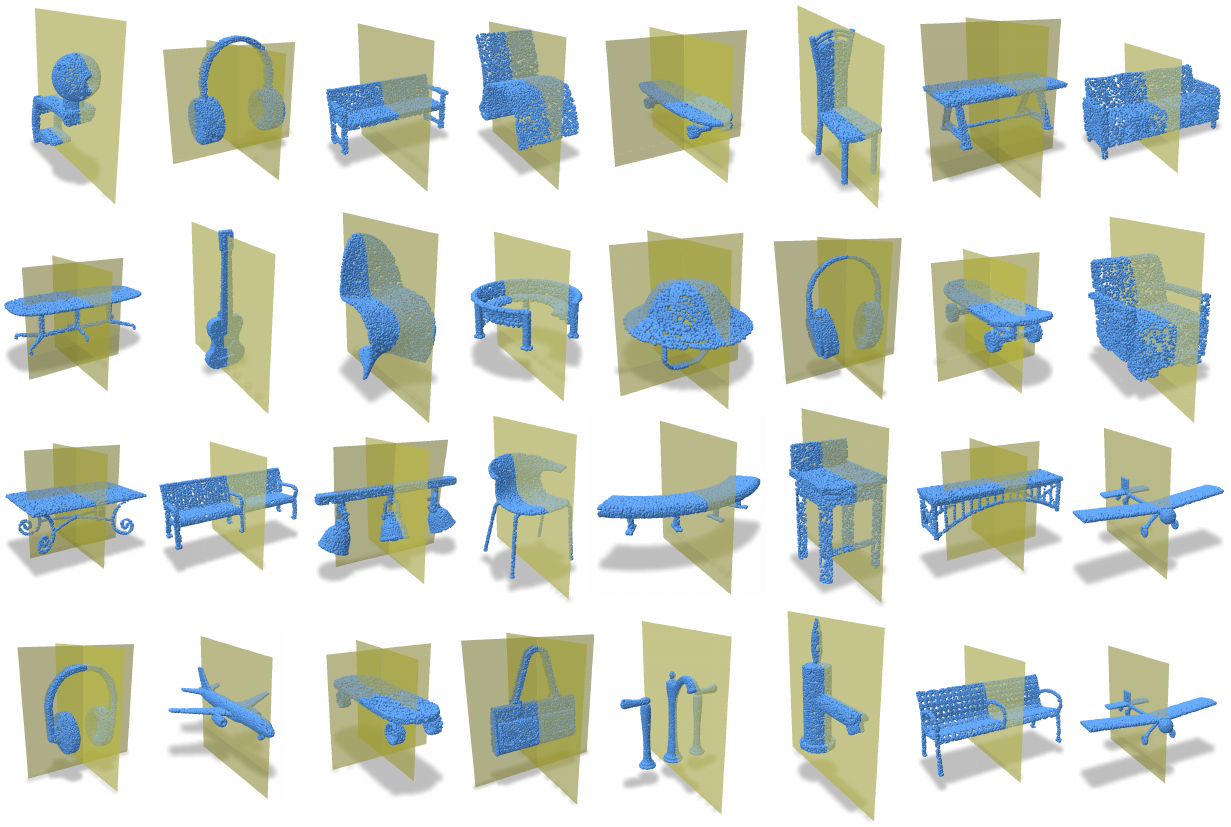}
\caption{Examples of symmetries detected with our method.}
\label{Fig:examples}
\end{figure*}

\paragraph{F-score.}
Given two planes $u=(u_x, u_y, u_z, u_d)$ and $v = (v_x, v_y, v_z, v_d)$ represented by their coefficients in plane equation, the distance between $u$ and $v$ is defined as $\delta_{uv} = \min(\|u-v\|,\|u+v\|)$, where $\|\cdot\|$ is the $L_2$ norm. The distance takes into account the case that for a symmetric plane, $u = -u$.

%\begin{equation}
%    \delta_{uv} = \min(\|u-v\|,\|u+v\|)
%\end{equation}

This metric leverages ground truth symmetry information. We adopt the F-score definition proposed in ~\cite{e3sym2023}. A detected plane is considered a \emph{true positive} (TP) if there exists a ground truth plane with a distance $\delta_{uv}$ below a specified threshold. Otherwise, it is classified as a \emph{false positive} (FP). Additionally, if a ground truth plane is not within the threshold distance of any predicted plane, it is deemed a \emph{false negative} (FN).

Given a certain threshold, the precision is defined as $PR=\frac{TP}{TP+FP}$, the recall is defined as $RE=\frac{TP}{TP+FN}$, and the F-score is defined as $\frac{2 PR\times RE}{PR + RE}$. To ensure a robust comparison, we compute the F-score using thresholds 0.05, 0.1, 0.15, and 0.2. The reported F-score the average between several thresholds.

\subsection{Evaluation}
We evaluated our method against E3Sym~\cite{e3sym2023} (the state-of-the-art method for symmetry detection) and Diffusion~\cite{Sipiran2014} (a training-free method that uses geometric features to detect symmetries). For the results with E3Sym, we downloaded the publicly available pre-trained model and conducted inferences on the test set. In this evaluation, the objects in the test set were arbitrarily rotated to assess the ability of the methods to detect symmetric structures in a general and robust manner. 

Table~\ref{tab:comparison} presents the results of our method compared to other approaches. Our method consistently outperforms the alternatives across both metrics, indicating its effectiveness and precision in detecting symmetries. Figure~\ref{Fig:examples} shows examples of symmetries detected with our method.

\begin{table}
    \centering
    \begin{tabular}{lrr}
        \toprule
        Method  & SDE ($\times 10^{-4}$) & F-score \\
        \midrule
        Diffusion~\cite{Sipiran2014}     & 7.10 & 0.46   \\
        E3Sym~\cite{e3sym2023}  & 6.30   & 0.69        \\
        \textbf{Ours} & \textbf{4.72}          & \textbf{0.70}        \\
        \bottomrule
    \end{tabular}
    \caption{Metrics F-score and SDE ($\times 10^{-4}$) evaluated on the ShapeNet dataset.}
    \label{tab:comparison}
\end{table}

\paragraph{Ablation on method's input.}
The objective of this ablation study is to demonstrate the effectiveness of the proposed features introduced in this article. We evaluated three variants of our method: (1) using only 3D coordinates as input to the neural network (without features), where the loss function is solely the Chamfer distance; (2) using uniform sampling on the sphere to generate the views; and (3) using Fibonacci sampling on the sphere to generate the views. The results are presented in Table~\ref{tab:ablation}.

\begin{table}
    \centering
    \begin{tabular}{lrr}
        \toprule
        Method  & SDE ($\times 10^{-4}$) & F-score \\
        \midrule
        No views (3D coords)     & 13.28 & 0.65   \\
        Views (Uniform sampling)  & 5.80   & 0.43 \\
        Views (Fibonacci sampling) & \textbf{4.72}          & \textbf{0.70}        \\
        \bottomrule
    \end{tabular}
    \caption{Metrics F-score and SDE ($\times 10^{-4}$) to compare the effectiveness of the input of our model.}
    \label{tab:ablation}
\end{table}

\begin{figure}[ht!]
\centering
\includegraphics[width=0.8\columnwidth]{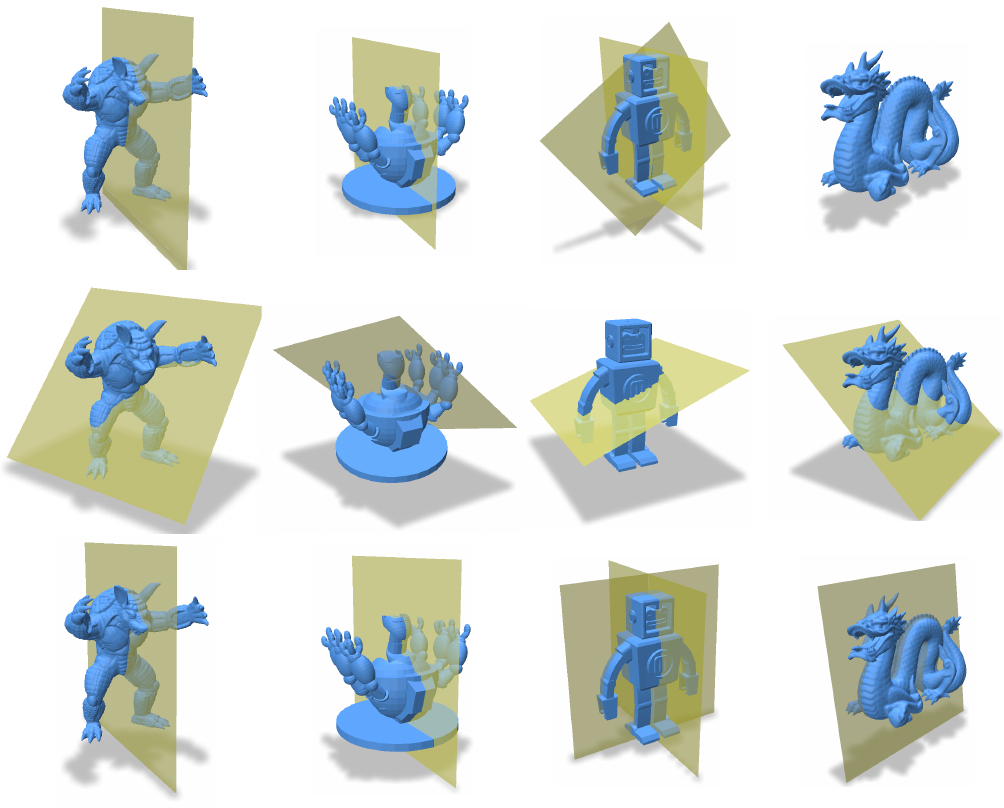}
\caption{Out-of-distribution evaluation. From top to bottom: detected symmetries from methods Diffusion, E3Sym, and our method. Note: Diffusion does not detect symmetries for the dragon.}
\label{Fig:out_distribution}
\end{figure}

\paragraph{Out of Distribution Analysis.}
We also aim to analyze the generalization capability of the compared methods. For this evaluation, we performed inference on objects with geometries significantly different from those in the ShapeNet dataset. The objective is to assess whether the methods can detect symmetries in novel geometries. The results are presented in Figure~\ref{Fig:out_distribution}.

Notably, our method consistently detects the symmetries of all tested objects. In contrast, E3Sym fails to identify the symmetries. We hypothesize that methods trained on specific datasets tend to learn patterns unique to the training data, thereby limiting their generalization ability. In contrast, our method focuses directly on the input object, enabling it to detect any observable symmetry regardless of prior training data.

\paragraph{Extended Analysis.}
We conducted an additional analysis using objects from the ShapeNet dataset. Specifically, we selected 1,600 objects from the airplane, car, chair, and table classes (400 objects per class). These classes were chosen because they are representative of the entire ShapeNet dataset but are underrepresented in the test set used in the previous section. Additionally, we only included objects with a symmetry plane whose normal is aligned with the positive X-axis.

We evaluated all three methods (Diffusion, E3Sym, and our proposed method) on this subset to obtain the predicted symmetry planes. For each object, we measured the angular error between the ground truth symmetry plane and the predicted planes. In cases where multiple symmetry planes were predicted, we selected the one with the smallest angular error. Finally, we computed the proportion of objects with predictions below a specified angular error threshold.

\begin{figure}[ht!]
\centering
\includegraphics[width=0.75\columnwidth]{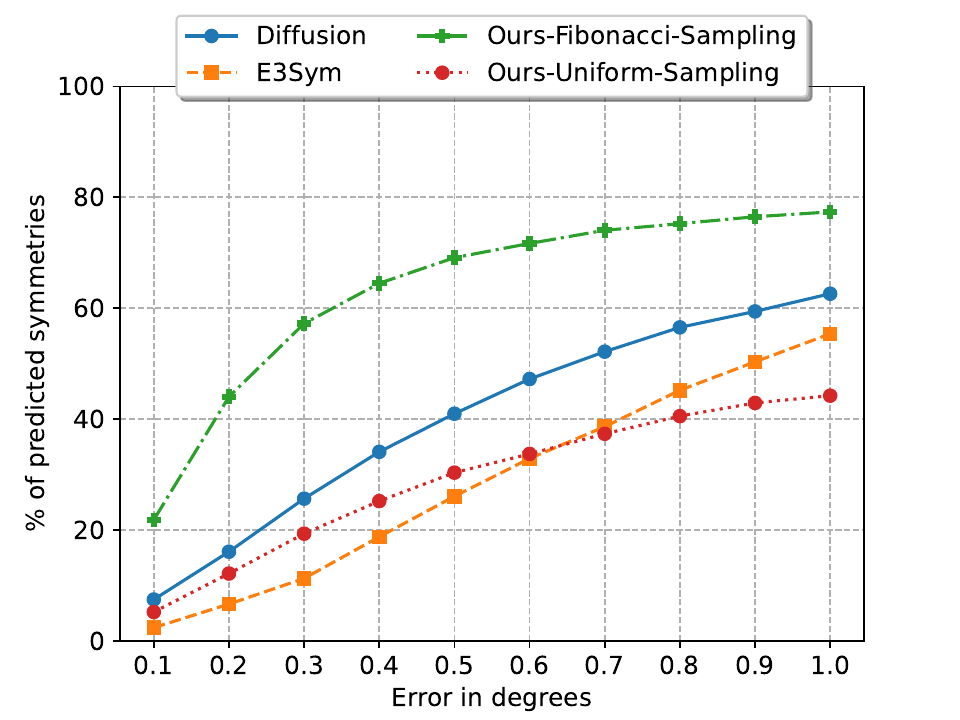}
\caption{Ratio of predicted symmetries with angular errors in interval $[0,1]$.}
\label{Fig:efficacy_1}
\end{figure}

Figure~\ref{Fig:efficacy_1} shows the proportion of predicted planes using different error thresholds ranging from $0°$ to $1°$ . Note that our method, when utilizing features from Fibonacci viewpoints, achieves the best performance for small error thresholds.

On the other hand, Figure 2 illustrates the proportion of predicted planes using error thresholds between $0°$ and $5°$ degrees. While E3Sym significantly increases the proportion of detected planes (outperforming our method) for thresholds above $2°$, it does so at the cost of allowing larger angular errors. However, it is important to note that our method achieves more than double the proportion of correctly predicted planes compared to E3Sym for a small angular threshold of $0.5°$.

This demonstrates that our method is both effective and precise in detecting 3D object symmetries.

\begin{figure}[ht!]
\centering
\includegraphics[width=0.75\columnwidth]{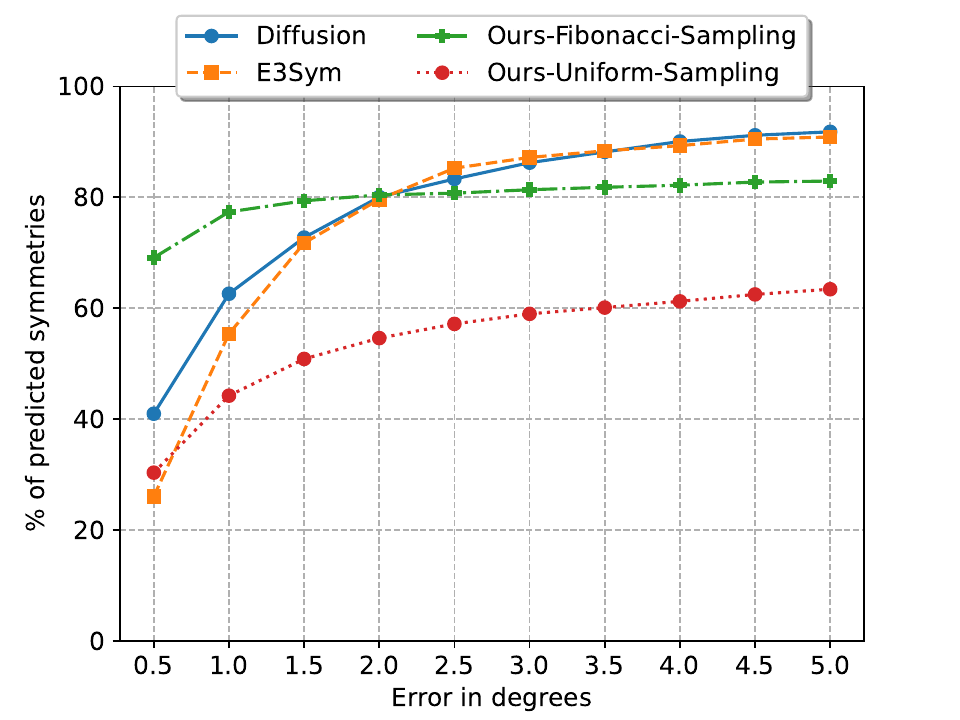}
\caption{Ratio of predicted symmetries with angular errors in interval $[0,5]$.}
\label{Fig:efficacy_5}
\end{figure}

\section{Conclusions}
\label{Sec:Conclusions}
In this paper, we introduced a self-supervised approach for detecting 3D reflectional symmetries, leveraging a self-prior framework that eliminates the need for large datasets and ground truth labels. We propose to compute visual features for 3D objects using foundational image models, which are useful to optimize a neural network trained with a symmetry-aware loss function. By focusing directly on the input object, our approach achieves remarkable generalization and robustness across diverse geometries.

Experimental results on the ShapeNet dataset demonstrate that our method outperforms state-of-the-art approaches in terms of both symmetry detection accuracy and generalization capability. Specifically, we show that the use of Fibonacci sampling for viewpoint generation and the integration of visual features significantly enhance performance, as evidenced by lower symmetry distance errors and higher F-scores. Furthermore, our ablation studies confirm the critical role of these design choices in improving detection precision.

In addition to in-distribution evaluations, we conducted out-of-distribution tests, revealing that our method consistently detects symmetries in novel geometries where other methods fail. This highlights the advantage of focusing on the intrinsic structure of the input object rather than relying on patterns learned from datasets.

Future work will explore extending our framework to detect rotational symmetries and handle objects with partial or occluded surfaces. Furthermore, integrating our approach with real-time applications in computer graphics and cultural heritage reconstruction remains a promising direction for practical deployment.
\bibliographystyle{acm}
\bibliography{ijcai25}

\end{document}